\documentclass[10pt,twocolumn,letterpaper]{article}

\usepackage{cvpr}              
\usepackage{makecell}
\usepackage{multirow}
\usepackage{booktabs}
\usepackage[table]{xcolor}
\usepackage{pifont}
\definecolor{cvprblue}{rgb}{0.21,0.49,0.74}
\usepackage[pagebackref,breaklinks,colorlinks,citecolor=cvprblue]{hyperref}

\def\paperID{*****} 
\def\confName{CVPR}
\def\confYear{2024}

\title{1st Place Solution for MeViS Track in CVPR 2024 PVUW Workshop: Motion Expression guided Video Segmentation}

\author{Mingqi Gao$^{1,4}$\footnotemark[2] \quad Jingnan Luo$^{2,}$\footnotemark[2] \quad Jinyu Yang$^{1,}$\footnotemark[1] \quad  Jungong Han$^{3,4}$ \quad  Feng Zheng$^{1,2,}$\footnotemark[1]\\
$^{1}$Tapall.ai \quad $^{2}$Southern University of Science and Technology \\ $^3$University of Sheffield \quad $^4$University of Warwick\\
\vspace{-0.5cm}
\\Team: Tapall.ai\\
{\tt\small \{mingqi.gao,jinyu.yang\}@tapall.ai, 12332444@mail.sustech.edu.cn}\\
{\tt\small jungonghan77@gmail.com, f.zheng@ieee.org}
}

\begin{document}
\maketitle
\renewcommand{\thefootnote}{\fnsymbol{footnote}}
\footnotetext[2]{Equal contributions.}
\footnotetext[1]{Corresponding authors.}
\renewcommand{\thefootnote}{\arabic{footnote}}
\begin{abstract}
Motion Expression guided Video Segmentation (MeViS), as an emerging task, poses many new challenges to the field of referring video object segmentation (RVOS). In this technical report, we investigated and validated the effectiveness of static-dominant data and frame sampling on this challenging setting. Our solution achieves a $\mathcal{J}\&\mathcal{F}$ score of 0.5447 in the competition phase and ranks 1st in the MeViS track of the PVUW Challenge. The code is available at: \href{https://github.com/Tapall-AI/MeViS_Track_Solution_2024}{https://github.com/Tapall-AI/MeViS\_Track\_Solution\_2024}. 
\end{abstract}    
\section{Introduction}
\label{sec:intro}

Pixel-level Video Understanding in the Wild (PVUW) is a workshop providing large-scale datasets, competitions, and discussions for video scene parsing, one of the fundamental problems in computer vision. Since 2021, PVUW has encouraged much improvement in video semantic segmentation and video panoptic segmentation. This year, two new subjects join PVUW: 1) Complex Video Object Segmentation (MOSE)~\cite{MOSE} and 2) Motion Expression guided Video Segmentation (MeViS)~\cite{ding2023mevis}, supplementing PVUW with object-centric pixel-level video understanding, vital to many real-world applications such as video editing and human-computer interactive systems. With large-scale videos, diverse/realistic challenges, and high-quality annotations, MOSE and MeViS build a competitive platform and encourage comprehensive and robust solutions. 

This technical report focuses on the MeViS subject: Motion Expression guided Video Segmentation, which aims to segment target objects in videos, guided by natural language expressions. Before MeViS, several benchmarks~\cite{gavrilyuk2018actor,khoreva2018video,seo2020urvos} have been proposed to encourage explorations in this field. Despite fostering surging research works, these benchmarks focus more on short videos with less same-category objects and static attributes (\eg, location and appearance). As a result, frame-level segmentation also predicts high-quality masks. This would weaken the investigation of temporal properties, vital to understanding real-world videos.

Recently, MeViS (Motion expressions Video Segmentation)~\cite{ding2023mevis} has been proposed to emphasise temporal properties in RVOS. Compared with previous benchmarks, MeViS brings several unique challenges: 1) Motion-dominant language expressions, 2) Complex scenes with multiple same-category instances, 3) One-to-more text-object pairs, and 4) Long videos. These challenges encourage RVOS to focus on dynamic attributes, comprehensive multi-modal interactions, and efficient long-term video understanding.  

The success of MTTR~\cite{botach2022end} and ReferFormer~\cite{wu2022language} motivates the community to consider transformer-based end-to-end architecture~\cite{carion2020end,ZhuSLLWD21} as the dominant paradigm. Given an input video and text, the paradigm encodes object queries from all frames and decodes text-relevant ones into masks. The difference between MTTR and ReferFormer lies in query encoding: The former encodes general object queries, and the latter considers language-guided queries, enlightening most following works. These could be roughly divided into two categories: 1) Robust vision-language alignments~\cite{li2023r2,li2023learning,miao2023spectrum}, which focus on the alignments between visual/textual properties; and 2) Temporal-aware interactions~\cite{han2023html,luo2023soc,yan2023referred}, where improvements leverage spatial and temporal properties to segment the targets. The more recent ones, SOC~\cite{luo2023soc} and MUTR~\cite{yan2023referred}, achieve SoTA performance on previous benchmarks due to their efficient interactions between object sequences and texts. This idea comes from VITA~\cite{heo2022vita}, a video instance segmentation method, which also inspired the latest RVOS SoTA: DsHMP~\cite{he2024decoupling}. Despite good results, they struggle with MeViS since they are trained on static-dominant data. In addition, MeViS consists of long videos, challenging RVOS in comprehensive video understanding and efficiency. 

With these new and realistic challenges, this report improves existing RVOS methods in training and inference schemes. Specifically, we consider MUTR~\cite{yan2023referred} as the baseline. With pre-trained weights on Ref-COCO series~\cite{yu2016modeling,mao2016generation} and Ref-YouTube-VOS~\cite{seo2020urvos}, we fine-tune them on MeViS. Masks with one-to-more text-object pairs are considered as a whole to encourage adaptive object perception based on texts. To balance comprehensive understanding and efficiency, we split long input videos into sub-videos via frame sampling. With these improvements, our solution ranks 1st in the MeViS Track. 

Experiments on the MeViS valid set indicate that previous RVOS data still contribute to this challenging setting due to their sufficient and well-aligned object masks and texts. In addition, ablations on sampling schemes reveal that there is much room for improvement in temporal modelling over long videos. Specifically, limited by computational resources, the temporal modules are trained with pseudo videos with less frames. During inference, however, videos have much more temporal contexts. This inconsistency leads to considering fewer frames (sampled) in temporal modules outperform the one with all frames. We hope these findings are helpful for future research. 

\section{Related Works}
\label{sec:relate}

This sections overviews representative methods and trends in referring and semi-supervised video object segmentation. 

\subsection{Referring Video Object Segmentation}
\label{relate:rvos}

Recent RVOS methods build their end-to-end architectures upon transformers. Specifically, given input videos and texts, the methods initialise fixed-number object queries to integrate vision-language contexts. Queries from different frames are considered a trajectory if they have the same index. The trajectory best matching with texts is decoded into masks on each frame. As pioneer works, MTTR~\cite{botach2022end} and ReferFormer~\cite{wu2022language} build foundation architectures with visual/textual encoders, multi-modal transformers, and mask decoders, motivating many following works. They improve RVOS in mainly two aspects: 1) Robust multi-modal alignments and 2) Temporal-aware interactions. 

With cyclic structural consensus, R2VOS~\cite{li2023r2} shows better results when text-referred objects are absent from frames. In SgMg~\cite{miao2023spectrum}, a spectrum-based multi-modal attention is proposed to improve query-guided mask predictions. FS-RVOS~\cite{li2023learning} improves RVOS to adapt new visual/textual concepts via the cross-modal affinity. As a versatile model, UNINEXT~\cite{yan2023universal} unifies different object perception tasks and data, generalising well on previous RVOS benchmarks. 

Previous methods rarely consider temporal properties or achieve this implicitly, \eg, with video-swin-transformer as backbone. In HTML~\cite{han2023html}, vision and language information are interacted over hierarchical temporal contexts. Motivated by VITA~\cite{heo2022vita}, which validates video understanding can be achieved via associating frame-level objects, recent methods encode temporal properties only from object queries, leading to end-to-end and efficient architectures: SOC~\cite{luo2023soc}, MUTR~\cite{yan2023referred}, and DsHMP~\cite{he2024decoupling}. The former two emphasise mutual multi-modal fusion and achieve SoTA performance on previous benchmarks, while the latter perform hierarchical multi-modal interactions and show high-quality results on ALL RVOS benchmarks. 

\subsection{Semi-supervised Video Object Segmentation}

Unlike RVOS, which specifies the target objects with texts, semi-supervised video object segmentation (SVOS) considers manually annotated masks (usually on the first frame) as targets~\cite{gao2023deep}. Therefore, SVOS methods focus on dense correspondence between frames and can propagate high-quality masks from one or several frames to the whole video. 

With this feature, most winner solutions~\cite{sun20221st,luo20241st} from previous RVOS competitions use SVOS methods to refine their results. In brief, they first select high-confident masks from overall predictions. Then, the masks are propagated to remaining frames to refine their corresponding results. The intuition behind the idea is that the offline RVOS methods struggle to generate spatial-temporal consistent object masks. This could be mitigated significantly via powerful SVOS methods (once the selected masks are high-quality). 

Memory-based paradigm (since STM~\cite{oh2019video}) has dominated SVOS due to its efficient, robust, and dense correspondence between frames. In particular, the paradigm considers not only the first frame annotations but also predictions from intermediate frames as references. This way, SVOS could better adapt to object changes. 

\begin{figure}[t]
\centering
\includegraphics[width=\linewidth]{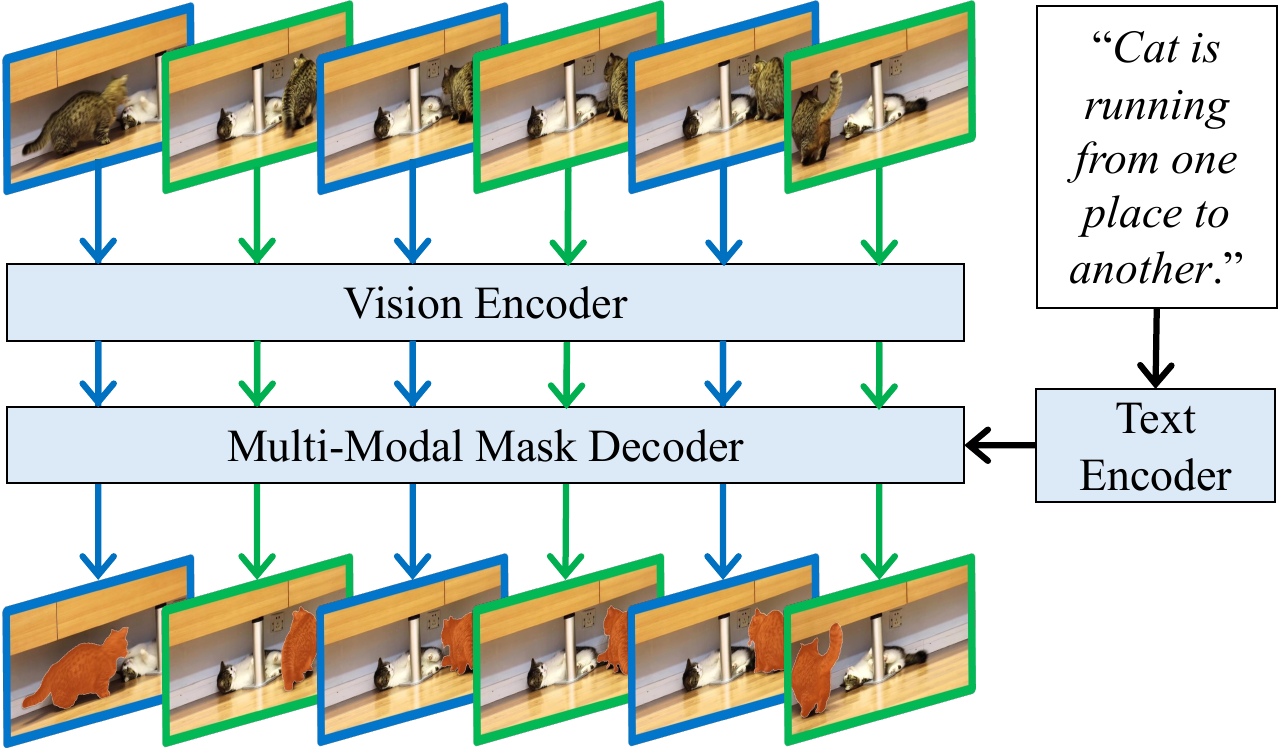}
   \caption{Overview of our solution. Given an input video, we divide all frames into $N$ subsets via non-continuous sampling. Here we take two subsets as an example. They are marked with \textcolor[RGB]{0,112,192}{Blue} and \textcolor[RGB]{0,176,80}{Green} boxes. In particular, each subset is segmented individually, guided by the input text, and combined for the final results. }
\label{fig:overview}
\end{figure}

Earlier SoTAs~\cite{seong2020kernelized,cheng2021rethinking} improve STM with robust cross-frame correspondence. The recent focus has been shifted to more challenging and realistic settings: long videos and complex scenes, motivating high-quality benchmarks~\cite{MOSE,hong2024lvos} and solutions. Specifically, XMem~\cite{cheng2022xmem} is proposed to segment long videos with dynamic memory management. AOT series~\cite{yang2021associating,yang2024scalable} consider object representations to enhance the robustness against complex scenes. By integrating object queries into dense correspondence, Cutie~\cite{cheng2024putting} significantly reduces the matching noise between frames and achieves the SoTA SVOS performance. 

\section{Method}
\label{sec:method}

Fig.~\ref{fig:overview} shows our solution, where we consider MUTR~\cite{yan2023referred} as the base model, with Swin-Transformer-Large as vision encoder and RoBERTa-base as text encoder. Given an input video with $T$ frames ($\mathcal{V}=\{v_t\in \mathbb{R}^{H\times W\times 3}\}^{T}_{t=1}$) and referring text $\mathcal{E}=\{e_i\}^L_{i=1}$ with $L$ words, we first sample $\mathcal{V}$ into $N$ subsets: $\{\mathcal{V}_n\}_{n=1}^N$. Then, we segment each subset individually under the guidance from $\mathcal{E}$, achieving mask subsets: $\{\mathcal{M}_n\}_{n=1}^N$. Finally, the masks are combined for the final predictions: $\mathcal{M}=\{m_t\in \mathbb{R}^{H\times W}\}_{t=1}^T$. 

\vspace{-0.4cm}
\paragraph{Training details.} With MUTR's weights jointly trained on Ref-COCO~\cite{yu2016modeling}, Ref-COCO+~\cite{yu2016modeling}, Ref-COCOg~\cite{mao2016generation}, and Ref-YouTube-VOS~\cite{seo2020urvos}, we perform fine-tuning on MeViS training videos. For the expressions specifying multiple objects, we consider all masks as a whole to encourage the model to perceive and segment all objects from videos. To better leverage pre-trained parameters, we follow MUTR to sample five frames as a pseudo video and use the same losses. The fine-tuning is performed for two epochs, where the learning rate is reduced to 10\% during the last one. 

\vspace{-0.4cm}
\paragraph{Inference details.} Given an input video, we resize each frame to keep its shorter size at 360. Unlike previous RVOS benchmarks, MeViS videos consist of much frames and thus cannot be inferred with one feed-forward pass. As diagrammed in Fig.~\ref{fig:overview}, we sample the video into $N=T\mid T_c$ subsets and perform referring segmentation individually. $T_c=30$ is the length of each subset. 

\vspace{-0.1cm}
\section{Experiments}
\label{sec:exp}

This section first shows our quantitative and qualitative results on the MeViS test set. Then, we provide ablations on MeViS validation set to show the solution's effectiveness and try to derive some insights for future research. 

\subsection{Main Results}

Tab.~\ref{tab:ranking} shows our quantitative results on the MeViS test set. 

\begin{table}[t]\renewcommand{\arraystretch}{1.01}
\small
\tabcolsep=0.2cm
  \centering
  \begin{tabular}{p{0.2\columnwidth}|p{0.2\columnwidth}<{\centering}p{0.2\columnwidth}<{\centering}p{0.2\columnwidth}<{\centering}}
  \toprule
   Team & $\mathcal{J}\&\mathcal{F}$ & $\mathcal{J}$ & $\mathcal{F}$ \\
  \noalign{\vspace{1.5pt}}
    \hline
    \noalign{\vspace{1.5pt}}
   Tapall.ai & \cellcolor{blue!17!white} 0.5447 (1) & 0.5048 (2) & \cellcolor{blue!17!white} 0.5846 (1) \\
   BBBiiinnn & 0.5420 (2) & \cellcolor{blue!17!white} 0.5097 (1) &	0.5743 (2) \\
    PPPPPsanG & 0.5280 (3) & 0.4853 (3) & 0.5707 (3) \\
    times & 0.5151 (4) & 0.4610 (4) & 0.5691 (4) \\
    Phan & 0.5075 (5) & 0.4562 (5) & 0.5588 (5) \\
  LIULINKAI & 0.4267 (6) & 0.3927 (6) & 0.4607 (6) \\
ntuLC & 0.3700 (7) & 0.3407 (7) & 0.3994 (7) \\

  \bottomrule
  \end{tabular}
  \caption{Quantitative results on the MeViS test set. }
  \label{tab:ranking}
\end{table}

\begin{table}[t]\renewcommand{\arraystretch}{1.01}
\small
\tabcolsep=0.15cm
  \centering
  \begin{tabular}{p{0.2\columnwidth}|p{0.2\columnwidth}<{\centering}|p{0.12\columnwidth}<{\centering}|p{0.12\columnwidth}<{\centering}|p{0.16\columnwidth}<{\centering}}
  \toprule
  Method & Backbone & Prev. & MeViS & $\mathcal{J}\&\mathcal{F}$\\
  \noalign{\vspace{1.5pt}}
    \hline
    \noalign{\vspace{1.5pt}}
    MUTR~\cite{yan2023referred} & Swin-L & \ding{51} & \ding{55} & 0.4343\\
    MUTR~\cite{yan2023referred} & Swin-L & \ding{51} & \ding{51} & \cellcolor{blue!17!white} 0.4857 \\
    SOC~\cite{luo2023soc} & V-Swin-B & \ding{51} & \ding{55} & 0.4394\\
    SOC~\cite{luo2023soc} & V-Swin-B & \ding{51} & \ding{51} & 0.4664 \\

  \bottomrule
  \end{tabular}
  \caption{Ablations on training data. `Prev.' indicates the use of Ref-COCO, Ref-COCO+, Ref-COCOg, and Ref-YouTube-VOS.}
  \label{tab:data}
\end{table}

\begin{table}[t]\renewcommand{\arraystretch}{1.01}
\small
\tabcolsep=0.069cm
  \centering
  \begin{tabular}{p{0.21\columnwidth}|p{0.112\columnwidth}<{\centering}p{0.112\columnwidth}<{\centering}p{0.112\columnwidth}<{\centering}p{0.112\columnwidth}<{\centering}p{0.112\columnwidth}<{\centering}p{0.112\columnwidth}<{\centering}}
  \toprule
   Method & 1 & 5 & 10 & 20 & 30 & 40 \\
  \noalign{\vspace{1.5pt}}
    \hline
    \noalign{\vspace{1.5pt}}
   N-continuous & 0.4725 & 0.4788 & 0.4811 & 0.4849 & \cellcolor{blue!17!white} 0.4857 & 0.4864 \\
   Continuous & 0.4725 & 0.4831 & 0.4855 & 0.4875 & 0.4864 & 0.4892 \\
   No sampling & 0.4607 & 0.4685 & 0.4730 & 0.4792 & 0.4822 & 0.4732 \\

  \bottomrule
  \end{tabular}
  \caption{Ablations on sampling methods. Five sub-video lengths are considered. The \colorbox{blue!17!white}{highlighted results} and the results evaluated in Tab.~\ref{tab:ranking} come from the same model. }
  \label{tab:clip}
\end{table}

\begin{figure}[t]
\centering
\includegraphics[width=\linewidth]{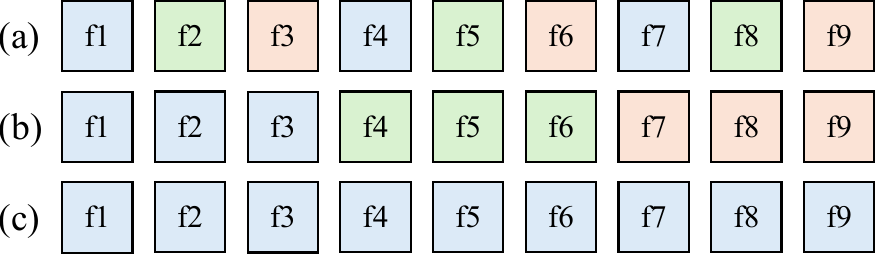}
   \caption{Difference between (a) Non-continuous sampling, (b) Continuous sampling, and (c) No sampling. Each box here denotes one frame, and the same colour boxes are sampled as pseudo videos for referring segmentation. Note that the best object trajectory selection is performed individually in each sampled video. For `No sampling', we still divide videos into subsets and predict masks upon those. During the selection, we feed all object queries into temporal modules and consider the resulting probabilities to select the best mask trajectory. }
\label{fig:clip}
\end{figure}

\vspace{-0.1cm}
\subsection{Ablations}

\begin{figure*}[t]
\centering
\includegraphics[width=.98\linewidth]{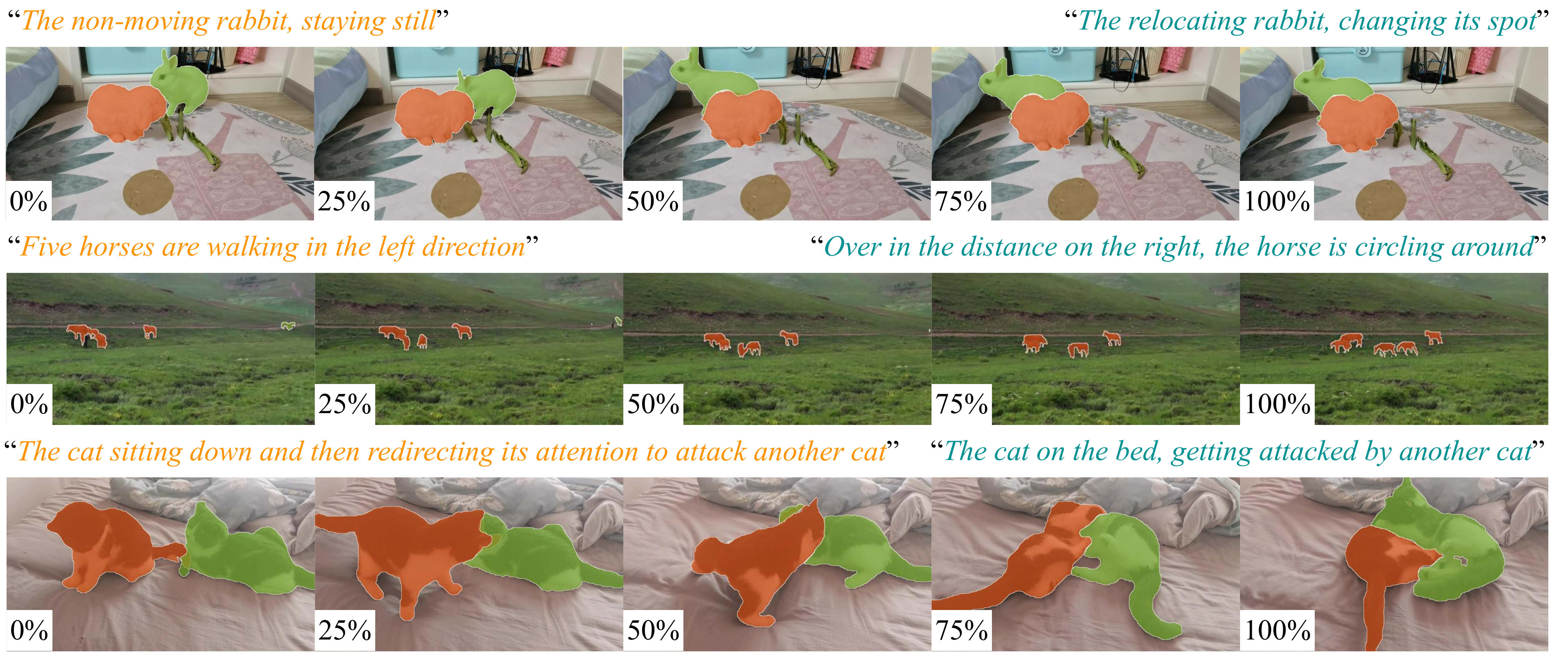}
\vspace{-0.2cm}
   \caption{Qualitative predictions of our solution on the MeViS valid set. \textcolor[RGB]{240,152,55}{Orange} and \textcolor[RGB]{63,143,145}{Green} masks are the predictions guided by the texts with the same colour. The percentage indicates the position of corresponding frame in the video. }
\label{fig:qualitative}
\end{figure*}

To validate the effectiveness of our solution, we show results in Fig.~\ref{fig:qualitative} and ablations on the MeViS valid set. 

\begin{figure*}[t!]
\centering
\includegraphics[width=.98\linewidth]{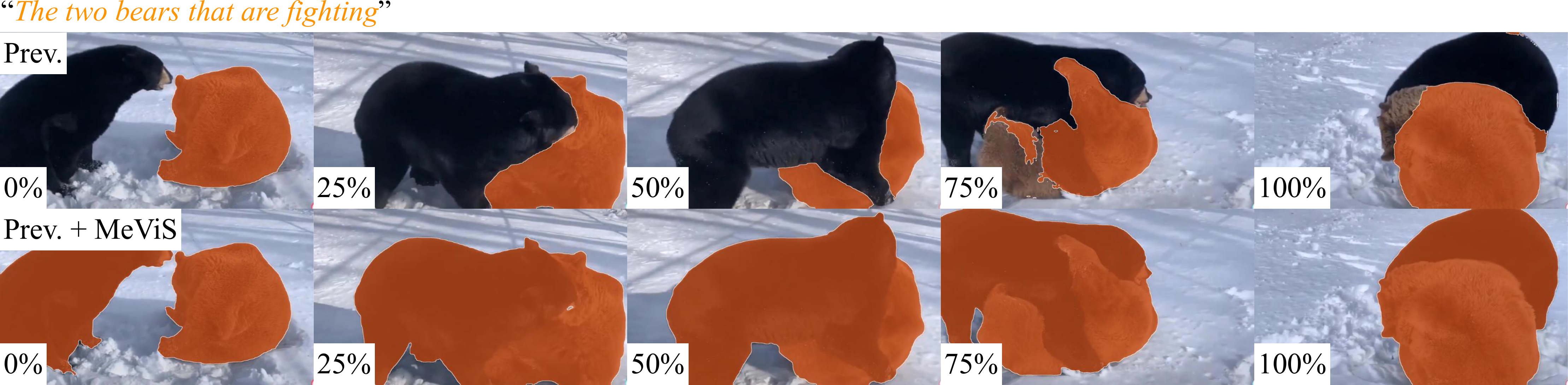}
\vspace{-0.2cm}
   \caption{Qualitative ablations on training data. The percentage indicates the position of corresponding frame in the video. }
\label{fig:qual_data}
\end{figure*}


\begin{figure*}[t!]
\centering
\includegraphics[width=.98\linewidth]{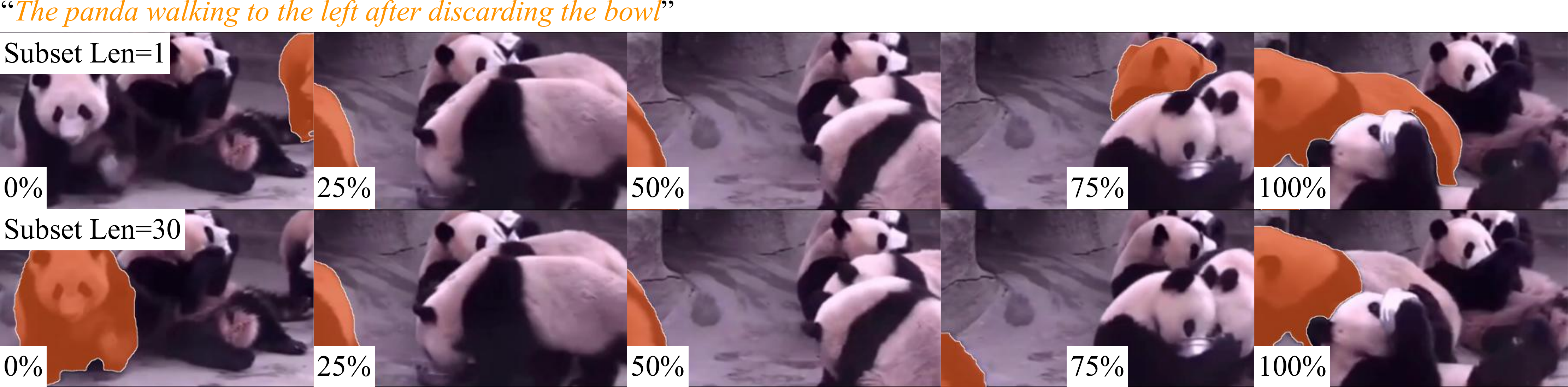}
\vspace{-0.2cm}
   \caption{Qualitative ablations on subset video length. The percentage indicates the position of corresponding frame in the video. }
\label{fig:length}
\end{figure*}

\vspace{-0.3cm}
\paragraph{Training method.}

Tab.~\ref{tab:data} compares $\mathcal{J}\&\mathcal{F}$ on different training data. To generalise the conclusion, we take another RVOS method with temporal properties (SOC~\cite{luo2023soc}) into account. MUTR and SOC share the same training and inference procedure. It is observed that the training data in previous benchmarks still contribute to this challenging setting, due to their sufficient and well-aligned object-text pairs. Fig.~\ref{fig:qual_data} shows that previous benchmarks enable RVOS methods to perceive objects. With MeViS and unified mask supervision, the methods work on more challenging applications with motion expressions or ones with plural nouns. 

\vspace{-0.3cm}
\paragraph{Sampling method.}

Tab.~\ref{tab:clip} ablates methods and hyper-parameters for sampling frames. The difference between these methods is diagrammed in Fig.~\ref{fig:clip}. Although temporal modules in MUTR enable us to collect and infer long-term object queries from videos, they are only trained with pseudo videos with five frames. The gap between training and inference temporal contexts struggles with temporal interactions over long videos. Results in Tab.~\ref{tab:clip} and Fig.~\ref{fig:length} show that temporal modules works better than frame-level predictions (sub-video length=1) but the performance cannot be improved further with more temporal contexts. 

\section{Conclusion}
\label{sec:conclusion}

This technical report explores the value of training data and temporal contexts for the challenging MeViS benchmark. The competitive results and ablations demonstrate that the well-aligned object-text data (even with primarily the static attributes) are helpful in motion expression-guided referring video segmentation. In addition, we investigate the effectiveness of temporal contexts and reveal room for improvement in the temporal multi-modal analysis of long videos.

{
    \small
    \bibliographystyle{ieeenat_fullname}
    \bibliography{main}
}


\end{document}